\documentclass{article}

\usepackage{graphicx} 
\usepackage{subfigure} 

\usepackage{natbib}

\usepackage{algorithm}
\usepackage{algorithmic}

\usepackage{hyperref}

\graphicspath{{../}}

\global\long\def\mednorm#1{\bar{#1}}

\global\long\def\firsttiernorm#1{\acute{#1}}

\global\long\def\secdtiernorm#1{\grave{#1}}

\title{A Rank Minrelation - Majrelation Coefficient}

\begin{document} 

\author{Patrick E. Meyer \\ \\ pmeyer@ulb.ac.be\\
Machine Learning Group\\
Computer Science Department\\
Universite Libre de Bruxelles\\
CP 212 - 1050 Brussels - Belgium}
\date{{\bf Technical Report} - April 2013}

\maketitle

\begin{abstract} 
Improving the detection of relevant variables using a new bivariate measure could importantly impact variable selection and large network inference methods.
In this paper, we propose a new statistical coefficient that we call
the \emph{rank minrelation coefficient}. We define a minrelation of
$X$ to $Y$ (or equivalently a majrelation of $Y$ to $X$) as a measure that estimate
 $p(X\leq Y)$ when $X$ and $Y$ are continuous random variables. The approach is similar to
Lin's concordance coefficient that rather focuses on estimating $p(X=Y)$. 
In other words, if a variable $X$ exhibits a minrelation to $Y$ then, as $X$ increases, $Y$
is likely to increases too. However, on the contrary to concordance or correlation,
the minrelation is not symmetric. More explicitly, if $X$ decreases,
little can be said on $Y$ values (except that the uncertainty on
$Y$ actually increases). In this paper, we formally define this new
kind of bivariate dependencies and propose a new statistical
coefficient in order to detect those dependencies. We show through
several key examples that this new coefficient has many interesting
properties in order to select relevant variables, in particular when
compared to correlation.
\end{abstract} 

\section{Introduction}
\label{intro}
When it comes to selecting relevant variables (relevance as defined in \citep{kojadinovic05}), many filter methods
try to find the subset of variables ($X_{S}$) that is the most predictive to the target variable ($Y$)  \citep{guyon03}. 
Some of these efficient filter methods (such as ranking \citep{duch03}, mRMR \citep{peng05}, FCBF \citep{yu04}) select a subset by identifying
relevant variables based only on bivariate measures. However, the
fact that the set $X_{S}$ is predictive to $Y$ do not ensure that
 its component variables $X_{i\in S}$ are relevant to
$Y$ \citep{meyer07fs}. Improving the detection of relevant variables using a bivariate measure could importantly impact  large network inference methods that rely heavily on 
bivariate measures  (such as correlation networks \citep{butte00} or Aracne \citep{margolin06}).
The objective of this paper is precisely to define a new statistical coefficient that improves the detection of relevant variables using only a bivariate measure. 
We will see that our new coefficient implicitly focuses on the little studied issue of heteroscedasticity. 
Let us first provide some examples:
\begin{enumerate}
\item When the price of aluminum increases the prices of cars is likely
to increase too. However, if the price of aluminum drops, the price
of cars might stay high because of other components or some technological
and economical considerations (i.e. other relevant variables). However,
if the price of cars become really low, then it is likely that the
price of their components, including aluminum, are considered as low
too (i.e. relatively to their average values). 
\item An increase in the level of adrenaline leads to an increased heart
rate, but a low level of adrenaline do not prevent a high heart rate (because of other relevant variables)
and a high heart rate do not mean a high adrenaline level. However,
a low level of adrenaline is likely to be observed in a person having
a low heart rate (w.r.t. to his/her usual heart rate). 
\item Let $Y=X_{1}.X_{2}$ with $X_{1}\in[0,1]$, and $X_{2}\in[0,1]$.
In such case, a low $X_{1}$ implies a low $Y$ but a high $X_{1}$
has little information on $Y$ (because a low $X_{2}$ automatically
means a low $Y$ whatever the value of $X_{1}$). However, a high
$Y$ (w.r.t. its average) automatically implies a high $X_{1}$. 
\end{enumerate}
In those examples, where variable dependencies are not quite correlations,
looking for correlations might be fastidious because in order to observe
joint variations, it might require that no other effect impacts the
measured variables or that all those effects cancel each other. 
It worth noting that the joint distribution we are focusing on, illustrated by example 3 and also Fig. (\ref{fig:Typical-implication}), has been identified in \citep{sahoo08} as the distribution observed in variables exhibiting an implication relationship (a probabilistic version of it).
However,  \citep{sahoo08} rely on discretization (in a maximum of three classes) to detect these dependencies whereas we will 
define a coefficient adapted to continuous variables, in Section \ref{sec:Minorelation}. Assumptions will be discussed in \ref{sec:Assumptions}. Properties of the new coefficient are stressed in Section\ref{sec:Properties}.  In section \ref{sec:Experiments}, preliminary experiments show the competitivity
of this new coefficient for variable selection w.r.t. correlation. 

\section{Minrelation \label{sec:Minorelation}}
In this section, we will define the minrelation of $X$ to $Y$ as an estimate $\hat p(X\leq Y)$ of $p(X\leq Y)$ for continuous random variables. \\
Respectively, the majrelation of $X$ to $Y$ will be defined as the estimate $\hat p(X\geq Y)$ of $p(X\geq Y)$. \\ 

\begin{figure}
\begin{centering}
\includegraphics[width=0.8\columnwidth]{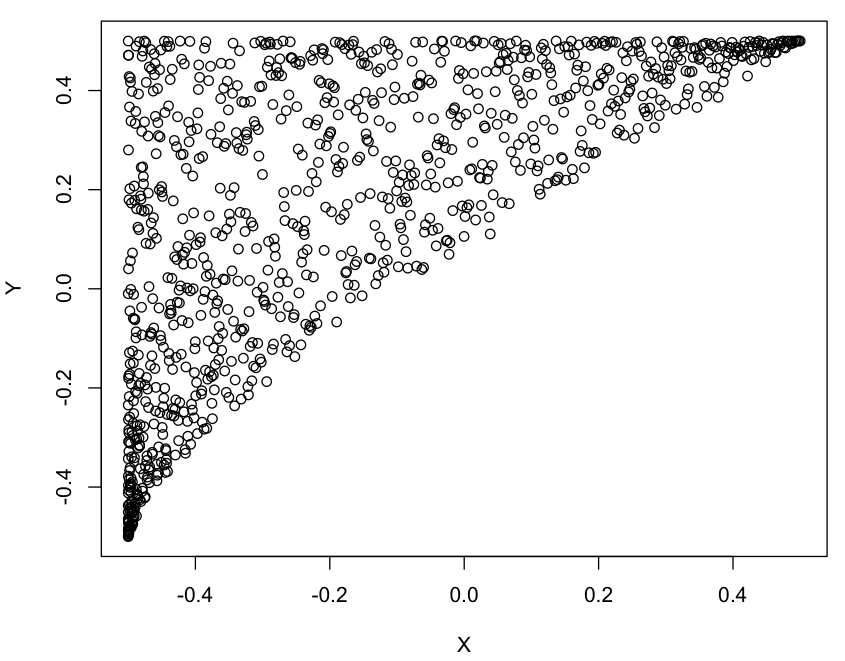}
\par\end{centering}
\caption{Typical plot of a minrelation between $X$ and $Y$. For a linear model, the variance of $Y|X$ decreases as $X$ increases and symmetrically the variance of $X|Y$ decreases as $Y$ decreases \label{fig:Typical-implication}}
\end{figure}

For example,  let $m$ samples of $X\in[-0.5,0.5]$ and $Y\in[-0.5,0.5]$ be drawn such that $\forall(x_{i},y_{i}),x_{i}\leq y_{i},i\in\{1,...,m\}$,
see Fig. (\ref{fig:Typical-implication}). 
In such case, we say there is a perfect minrelation of $X$ to $Y$
(or equivalently a perfect majrelation of $Y$ to $X$), i.e. $\hat p(X\leq Y)=1$.
In a linear model, the variance of $Y|X$ would decrease as $X$ increases and symmetrically, the variance of $X|Y$ decreases as $Y$ decreases. In other words, heteroscedasticity is a major component of a minrelation.

Let us now relax the constraint and assume that $\exists j\in\{1,...,m\},x_{j}>y_{j}$. In that case,
a simple way to estimate $p(X\leq Y)$ is to count the sample points $i$ that are concordant ($C$) with $x_i \leq y_i$,
$$\hat p(X\leq Y)=\frac{\sum_{i}I(x_{i}\leq y_{i})}{m}$$ with $I$ being the indicator function.
Similarly, we can count the sample points that are discordant ($D$) with it: 
$$\hat  p(X\leq Y)=1-\frac{\sum_{i}I(x_{i}>y_{i})}{m}$$

In order to create a coefficient that range between -1 and +1, we can compute
\begin{equation}
\frac{C-D}{C+D}=\frac{\sum_{i}I(x_{i}\leq y_{i})-\sum_{i}I(x_{i}>y_{i})}{m}
\label{eq:minrelation0}
\end{equation}

The formula above focuses on the trade-off between $\hat p(X\leq Y)$ 
and $\hat p(Y\leq X)$. The problem with the proposed
coefficient lies in the case where the joint distribution is symmetric.
Ideally a high concordance (i.e. $\hat p(X=Y)$ close to 1) should lead to a high minrelation of $X$ to $Y$ (i.e. $\hat p(X\leq Y)$ close to 1) together
with a high minrelation of $Y$ to $X$ (i.e. $\hat p(Y\leq X)$ close to 1). However, in this
case, highly concordant variables and independent variables would
both lead to a minrelation coefficient close to zero. 

Note that there are four possible minrelations/majrelations
\begin{enumerate}
\item $\hat p(X\leq Y)$ equivalently $\hat p (-Y\leq-X)$
\item $\hat p(X\leq-Y)$ equivalently $\hat p(Y\leq-X)$
\item $\hat p(-X\leq Y)$ equivalently $\hat p(-Y\leq X)$
\item $\hat p(-X\leq-Y)$ equivalently $\hat p(Y\leq X)$
\end{enumerate}
As a result, we rather propose another coefficient, that we denote by
$\iota(X,Y)$, and which focuses on the trade-off between $\hat p(X\leq Y)$
and $\hat p(X\leq-Y)$, i.e.,
\begin{small}
\begin{equation}
\iota(X,Y)=\frac{\sum_{i}I(x_{i}<-y_{i})-\sum_{i}I(x_{i}>y_{i})}{\sum_{i}I(x_{i}<-y_{i})+\sum_{i}I(x_{i}>y_{i})}\label{eq:minrelation1}
\end{equation}
\end{small}

\begin{figure}
\begin{centering}
\includegraphics[width=0.8\columnwidth]{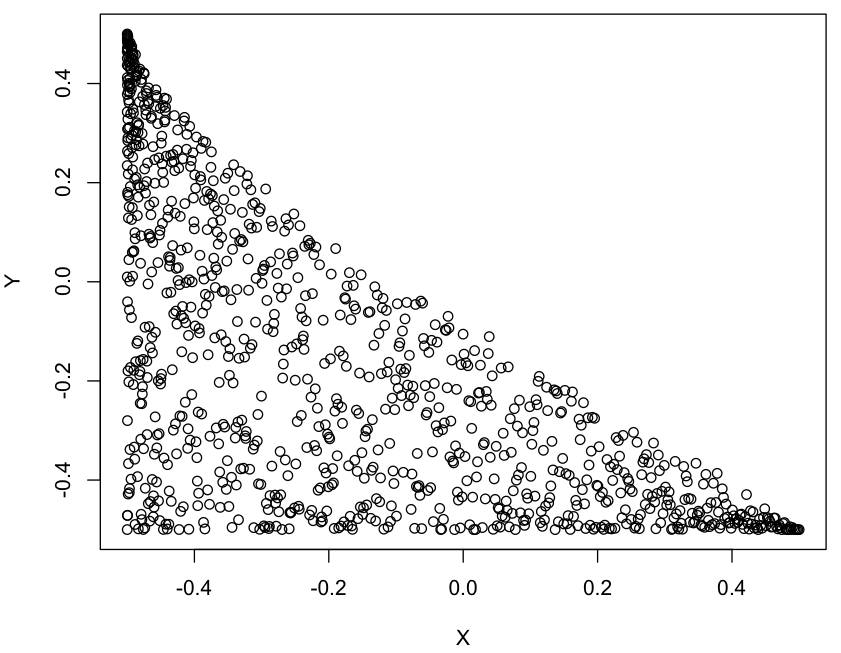}
\par\end{centering}
\caption{Typical plot of a minrelation between $X$ and $-Y$. \label{fig:Typical-implication2}}
\end{figure}

As a result, when the distribution is close to Fig. (\ref{fig:Typical-implication}), $\iota$ is close to +1. However, when the distribution is close to Fig. (\ref{fig:Typical-implication2}), $\iota$ is close to -1.

Although formula (\ref{eq:minrelation1}) can be applied to continuous random variables, it penalizes equally a point close to the diagonal $y=x$
and one far from it. Not only the agreement between which side of the diagonal falls each
realization $(x_{i},y_{i})$, but the agreement between their relative
distance to the diagonal, should matter. Hence, similarly to the concordance
coefficient, we adopt here squared distance to the diagonal \citep{lin89}.

Hence, the higher the sum of the squared distances of the
data points $i$ in discordance with the relation ($x_i\leq y_i$), the lower $\hat p(X\leq Y$).
As a result, iota becomes
\begin{equation}
\frac{\sum_{i}I(x_{i}<-y_{i})(x_{i}+y_{i})^{2}-\sum_{i}I(x_{i}>y_{i})(x_{i}-y_{i})^{2}}{\sum_{i}I(x_{i}<-y_{i})(x_{i}+y_{i})^{2}+\sum_{i}I(x_{i}>y_{i})(x_{i}-y_{i})^{2}}\label{eq:minrelation}
\end{equation}

Although this formula looks less simple than the concordance coefficient, its
computation is as straightforward.

\section{Assumptions\label{sec:Assumptions}}

We have implicitly assumed that $X$ and $Y$ are centered and normalized, but
the question that arise now is: \emph{how should we center and normalize those variables?} 
In other words, how can we compare variables
with very different ranges of values (such as aluminum prices and
car prices or adrenaline level and heart rate,...)? 

The concordance coefficient applied to centered and normalized variables boils down to Pearson's correlation.
One strategy that has been adopted for normalizing variables in correlations consists in mapping
marginals to ranks. For example, let 
\begin{equation}
\mednorm X=\frac{r(X)}{m}
\end{equation}
\begin{equation}
\mednorm Y=\frac{r(Y)}{m}
\end{equation}
with $r$ the function that return the rank (in increasing order)
of $x_{i}$ (resp. $y_{i}$) in the samples of $X$ (resp. in $Y$).
In such case, where data are converted into ranks, the marginals are
uniform distributions. As a result, when the concordance coefficient
is applied on variable that have been first converted to ranks, it fits 
 \emph{Spearman's rank correlation.}

However, a minrelation automatically implies that at least one marginal
is asymmetric (unless in the particular case of a correlation). Obviously
applying the minrelation coefficient on uniform distributions such
as rank data will lead to wrong results because it is impossible to
have $\forall i\in\{1,...,m\},\ r(x_{i})\leq r(y_{i})$
unless $\forall i\in\{1,...,m\},\ r(x_{i})=r(y_{i})$. Table (\ref{tab:minorelation-binary})
illustrates this phenomenon for binary variables where $p(x_0,y_1)>0$ automatically implies non-symmetrical marginals. In fact, both in
Fig. (\ref{fig:Typical-implication}) and in Table (\ref{tab:minorelation-binary}),
we can see that the distribution of $X$ is a decreasing triangular distribution and that the distribution of 
$Y$ is increasing triangular. 

\begin{center}
\begin{table}
\begin{centering}
\begin{tabular}{|c||c|c||c|}
\hline 
$p(X\leq Y)$ & $x_{0}$ & $x_{1}$ & $p(Y)$\tabularnewline
\hline 
\hline 
$y_{0}$ & 0.33 & 0 & 0.33\tabularnewline
\hline 
$y_{1}$ & 0.33 & 0.33 & 0.66\tabularnewline
\hline 
\hline 
$p(X)$ & 0.66 & 0.33 & 1\tabularnewline
\hline 
\end{tabular}%

\begin{tabular}{|c||c|c||c|}
\hline 
$p(X\leq-Y)$ & $x_{0}$ & $x_{1}$ & $p(Y)$\tabularnewline
\hline 
\hline 
$y_{0}$ & 0.33 & 0.33 & 0.66\tabularnewline
\hline 
$y_{1}$ & 0.33 & 0 & 0.33\tabularnewline
\hline 
\hline 
$p(X)$ & 0.66 & 0.33 & 1\tabularnewline
\hline 
\end{tabular}
\par\end{centering}

\caption{Examples of minrelation between binary variables. The first table is analogous to Fig. (\ref{fig:Typical-implication})  and the second one is analogous to Fig. (\ref{fig:Typical-implication2}).  \label{tab:minorelation-binary}}
\end{table}

\par\end{center}

Fortunately, we can map the data to a decreasing triangular distribution
by weighting linearly the uniform distribution. In such case, the smallest
rank should have the lowest weight and the highest rank should have
the highest. In other words, the weight $w_{i}$ of each sample is
precisely its ranking $r(x_{i})$. 

This leads to 

\begin{equation}
\secdtiernorm X=\frac{r(X)^{2}}{m^{2}}-0.5
\end{equation}
in order to have a centered decreasing triangular distribution, and

\begin{equation}
\firsttiernorm Y=0.5-\frac{r(-Y)^{2}}{m^{2}}
\end{equation}
in order to obtain an increasing triangular distribution (that mirrors
$X$). Note that $r(-Y)$ is the ranks of $Y$ in decreasing order
(instead of the increasing one). 

This transformation not only map the range of values of $X$ and $Y$
to the interval $[-0.5,0.5]$ but also map $E[\secdtiernorm X]$ to $-0.1666$
and $E[\firsttiernorm Y]$ to 0.1666 that are precisely the values
 observed in the symmetric joint distribution
illustrated in Table (\ref{tab:minorelation-binary}) and also in Fig. (\ref{fig:Typical-implication}). 

If we consider the estimation of $\hat p(X\leq-Y)$ (instead of $\hat p(X\leq Y)$,
see Table (\ref{tab:minorelation-binary})), we should rather use $\secdtiernorm Y$  instead of $\firsttiernorm Y$, in order to have a decreasing triangular
distribution instead of an increasing one, with
\begin{equation}
\secdtiernorm Y=\frac{r(Y)^{2}}{m^{2}}-0.5
\end{equation}

Hence, we can define $\iota(X,Y)$, the\emph{ rank minrelation coefficient},
as the\emph{ minrelation coefficient}  but applied on the variables
converted to increasing (resp. decreasing) squared ranks. This is
done by plugging $\secdtiernorm X$, $\firsttiernorm Y$ and $\secdtiernorm Y$
in eq. \ref{eq:minrelation}. The rank minrelation coefficient $\iota(X,Y)$
becomes
\begin{equation}
\frac{\sum_{i}I(-\secdtiernorm x_{i}<\secdtiernorm y_{i})(\secdtiernorm x_{i}+\secdtiernorm y_{i})^{2}-\sum_{i}I(\secdtiernorm x_{i}>\firsttiernorm y_{i})(\secdtiernorm x_{i}-\firsttiernorm y_{i})^{2}}{\sum_{i}I(-\secdtiernorm x_{i}<\secdtiernorm y_{i})(\secdtiernorm x_{i}+\secdtiernorm y_{i})^{2}+\sum_{i}I(\secdtiernorm x_{i}>\firsttiernorm y_{i})(\secdtiernorm x_{i}-\firsttiernorm y_{i})^{2}}
\end{equation}

\section{Properties\label{sec:Properties}}

This new coefficient benefits from the following properties
\begin{enumerate}
\item $-1\leq\iota(X,Y)\leq1$
\item if $\forall(x_{i},y_{i}),x_{i}\leq y_{i}$ then $\iota(X,Y)=1$ (similarly
if $\forall(x_{i},y_{i}),y_{i}\leq x_{i}$ then $\iota(Y,X)=1$)
\item if $X$ and $Y$ are independent, then $\iota(X,Y)=\iota(Y,X)=0$ 
\item if $\rho(X,Y)=1$ then $\iota(X,Y)=1$ and $\iota(Y,X)=1$ 
\item if $\rho(X,Y)=-1$ then $\iota(X,Y)=-1$ and $\iota(Y,X)=-1$ 
\end{enumerate}

Hence,  high correlation implies minrelated variables but high minrelations can happen with poorly correlated variables.\\
Moreover, thanks to the squared ranks conversion discussed earlier the joint distribution of $X$ and $Y$ becomes symmetric w.r.t.
the diagonal $y=-x$, which leads to $\iota(X,Y)\simeq \iota(-Y,-X)$.\\
Obviously, there is a similar coefficient benefitting from the same properties: $\iota_{2}(X,Y)$ focusing
on the trade-off between $\hat p(X\leq Y)$ and $\hat p(-X\leq Y)$ (instead of $\hat p(X\leq -Y)$).

In fact, $\iota(X,Y)=\iota_{2}(-Y,-X)$ and reciprocally $\iota_{2}(X,Y)=\iota(-Y,-X)$. Hence, $\iota(X,Y)$ and $\iota_{2}(X,Y)$ return close values once data have been converted to squared ranks.

However, if $X$ and $Y$ joint distribution is not symmetric w.r.t. the diagonal
$y=-x$, these coefficients would actually answer different questions. For
example, let us assume that increasing the dosage of a medication
increases the probability of getting cured, i.e. $p(high\ dose,cured)$
is close to 1. In such case, $\iota(dose,cure)$ sort of compares
$p(high\ dose,cured)$ with $\hat p(high\ dose,not\ cured)$ whereas $\iota_{2}(dose,cure)$
sort of compares $p(high\ dose,cured)$ with $\hat p(low\ dose,cured)$
(and hopefully $\iota_{2}<\iota$, being cured with a low dose is
more likely than not being cured with a high dose).
 
\section{Experiments\label{sec:Experiments}}

The goal of this section is to show the usefulness of the new rank
minrelation coefficient $\iota$ in data analysis (when compared to
Spearman's $\rho$). We first demonstrate $\iota$ competitiveness
on toy examples, then on artificial and real datasets by plugging
it into the well-known ranking variable selection method (because
it is based on pairwise similarity measure).

\subsection{Multiplication}

As mentioned in the introduction, one of the easiest way of generating
minrelation consists in defining $B$ and $C$ as independent and
uniformly distributed variables (in the positive range) and let $A=B.C$
because it results that $\hat p(A\leq B)=1$ and $\hat p(A\leq C)=1$ (after $A,B$
and $C$ are converted to squared ranks). We report in Table (\ref{tab:toyex1})
the average value of minrelation and correlation coefficients over
1000 runs of the above setting where each variable is constituted
of 1000 samples.

\begin{table}
\begin{centering}
\begin{tabular}{|c|c|c|c|}
\hline 
$X$ & $A$ & $A$ & $B$\tabularnewline
\hline 
$Y$ & $B$ & $C$ & $C$\tabularnewline
\hline 
\hline 
$\rho(X,Y)$ & 0.66 & 0.66 & 0.00\tabularnewline
\hline 
$\iota(X,Y)=-\iota(X,-Y)$ & 0.99 & 0.99 & 0.00\tabularnewline
\hline 
$\iota(-Y,X)=-\iota(-Y,-X)$ & -0.99 & -0.99 & 0.00\tabularnewline
\hline 
$\iota(-X,Y)=-\iota(-X,-Y)$ & -0.79 & -0.79 & 0.00\tabularnewline
\hline 
$\iota(Y,X)=-\iota(Y,-X)$ & 0.77 & 0.77 & 0.00\tabularnewline
\hline 
\end{tabular} 
\par\end{centering}

\caption{Correlation coefficient $\rho$ and minrelation coefficient $\iota$
where ($A,B$) and ($A,C$) have minrelation dependencies and ($B,C$)
are independent (results are averaged over 1000 repetitions). \label{tab:toyex1}}
\end{table}

We observe that $\iota(X,Y)\simeq\iota(-Y,-X)$ and $\iota(Y,X)\simeq\iota(-X,-Y)$
exhibit quite close values as it is expected with the mapping of samples
to squared ranks.

\subsection{Linear dependencies}

Let's take another toy example with variables having linear dependencies.
Let $A=3B+2C+D$ with $B$, $C$ and $D$ independent and normally
distributed variables $N(0,1)$. The averaged results of 1000 runs
with each variables having 1000 samples are reported in Table (\ref{tab:toyex2}).

\begin{table}
\begin{centering}
\begin{tabular}{|c|c|c|c|}
\hline 
$X$ & $A$ & $A$ & $A$\tabularnewline
\hline 
$Y$ & $B$ & $C$ & $D$\tabularnewline
\hline 
\hline 
$\rho(X,Y)$ & 0.79 & 0.52 & 0.26\tabularnewline
\hline 
$\iota(X,Y)$ & 0.98 & 0.81 & 0.46\tabularnewline
\hline 
$\iota(-Y,X)$ & -0.98 & -0.81 & -0.46\tabularnewline
\hline 
$\iota(-X,Y)$ & -0.98 & -0.81 & -0.46\tabularnewline
\hline 
$\iota(Y,X)$ & 0.98 & 0.81 & 0.46\tabularnewline
\hline 
\end{tabular}
\par\end{centering}

\caption{Correlation coefficient $\rho$ and minrelation coefficient $\iota$
where ($A,B$), ($A,C$) and ($A,D$) are all having a linear dependency
(results are averaged over 1000 repetitions). \label{tab:toyex2}}
\end{table}

As expected, when the dependency between two variables is symmetric
(i.e. in a linear setting), we observe close values for $\iota(X,Y)\simeq\iota(Y,X)$.

\subsection{Multiplication together with linear dependencies}

Let 
\begin{equation}
\begin{array}{c}
G=A+E\\
A=BCD
\end{array}
\label{eq:toyex3}
\end{equation}
and observe (in Table (\ref{tab:toyex3})) which of $G$ and the independent
and uniformly distributed variables $B,C,D$ are more relevant in
order to predict $A$ using both correlation and minrelation coefficients
knowing that $E$ is a normally distributed variable $N(0,0.15)$
denoting the noise. 

\begin{table}
\begin{centering}
\begin{tabular}{|c|c|c|c|c|c|}
\hline 
$X$ & $A$ & $A$ & $A$ & $A$ & $A$\tabularnewline
\hline 
$Y$ & $B$ & $C$ & $D$ & $E$ & $G$\tabularnewline
\hline 
\hline 
$\rho(X,Y)$ & 0.53 & 0.53 & 0.53 & 0.00 & 0.57\tabularnewline
\hline 
$\rho(-X,Y)$ & -0.53 & -0.53 & -0.53 & 0.00 & -0.57\tabularnewline
\hline 
\hline 
$\iota(X,Y)$ & 0.97 & 0.97 & 0.97 & 0.00 & 0.92\tabularnewline
\hline 
$\iota(-Y,X)$ & -0.98 & -0.98 & -0.98 & 0.00 & -0.87\tabularnewline
\hline 
$\iota(-X,Y)$ & -0.69 & -0.69 & -0.69 & 0.00 & -0.78\tabularnewline
\hline 
$\iota(Y,X)$ & 0.64 & 0.64 & 0.64 & 0.00 & 0.85\tabularnewline
\hline 
\end{tabular}
\par\end{centering}

\caption{Correlation coefficient $\rho$ and minrelation coefficient $\iota$
averaged over 1000 repetitions of the example given in Eq. (\ref{eq:toyex3}).\label{tab:toyex3}}
\end{table}

Interestingly $\rho$ would rank variable $G$ first (higher correlation
with $A$) whereas $\iota(X,Y)$ would rank it after $B,C,D$. However,
if we look at $\iota(-X,Y)$ we can observe the same ranking than
for $\rho$. The minrelation coefficient kind of splits the correlation
signal into two parts: one coming from $(x_i \leq y_i)$ and another
coming from $(x_i \leq-y_i)$. Hence, in the following, when we are not
interested by the directionality of the minrelation, but rather by
the ordering of relevance provided by each criterion (correlation
vs minrelation), we will report $\rho^{2}$ and $\max\iota^{2}$ the
maximum over the four possible values of $\iota$ (squared in order
to avoid negative values).

\subsection{Artificial dataset}

The question that arise at this point is: \emph{is $\max\iota^{2}$
able to discriminate between relevant and irrelevant variables better
than $\rho^{2}$?} In order to answer this question, we make use of
a synthetically generated dataset where relevant and irrelevant variables
are known.  In this experiment, we compare the performances of variables
ranking by using $\rho^{2}$ and $\max\iota^{2}$ on our artificial
datasets. We consider a ranking strategy superior if the average
position of the relevant variables using that criterion is lower than
for the other criterion. The rationale being that a better selection
criterion should return a lower average position (i.e. relevant variables
should be ranked first). 

As artificial datasets, we adopt the 10 datasets of 100 variables
coming from the DREAM4 challenge (i.e. KO1...KO5, MF1...MF5) where
the goal was to identify predictor variables for each variables of
the dataset (it is a network inference task) \citep{marbach09}. However,
in this case, we focus only on the few variables per dataset that
have more than 10 predictors. This minimal number of predictors ensure
some stability in the results reported. Indeed, if one predictor variable
happens to be badly ranked by chance there are at least 9 others that
could compensate (if the criterion is indeed superior). The number
of target variables (i.e. ranking tasks), wins and losses of each
criterion for each dataset are reported in Table (\ref{tab:artifdata}).

\begin{table}
\begin{centering}
\begin{tabular}{|c|c|c|c|}
\hline 
DATASET & targets & rank-$\rho^{2}$ wins & rank-$\iota^{2}$ wins\tabularnewline
\hline 
\hline 
KO1 & 5 & 0 & 5\tabularnewline
\hline 
KO2 & 6 & 2 & 4\tabularnewline
\hline 
KO3 & 3 & 0 & 3\tabularnewline
\hline 
KO4 & 5 & 2 & 3\tabularnewline
\hline 
KO5 & 6 & 2 & 4\tabularnewline
\hline 
MF1 & 5 & 2 & 3\tabularnewline
\hline 
MF2 & 6 & 2 & 4\tabularnewline
\hline 
MF3 & 3 & 0 & 3\tabularnewline
\hline 
MF4 & 5 & 2 & 3\tabularnewline
\hline 
MF5 & 6 & 3 & 3\tabularnewline
\hline 
\hline 
Tot & 50 & 15 & 35\tabularnewline
\hline 
\end{tabular}
\par\end{centering}

\caption{Correlation coefficient $\rho^{2}$ vs minrelation coefficient $\max\iota^{2}$
in ranking strategies on target variables having more than 10 predictors
in the 10 datasets of the DREAM4 competition. Column 1 indicates the
dataset, column 2 indicates the number of variables having more than
10 predictors in that dataset and columns 3 and 4 reports the wins
and losses of the two ranking methods on those target variables. A
method wins if the average position of the predictors in the ranking
is lower than for the other method.\label{tab:artifdata}}
\end{table}

We observe that $\max\iota^{2}$ exhibits results that are more than
twice better than $\rho^{2}$ on a task that consists in identifying
the known set of predictors of target variables in ten artificial
datasets.

\subsection{Real datasets }

In the previous task, the variable to be selected were known in advance.
It is usually not the case in real datasets. In order to compare $\rho$
and $\iota$ coefficients on real data, we evaluate the prediction accuracy
of different learning algorithms (i.e. linear model, random forest
and radial SVM) using as input variables the best ranked ones using
$\rho^{2}$ and $\max\iota^{2}$. We assume here that a better criterion
leads to a better ranking of variables which in turn leads to better
prediction performances of a model built on these top ranked variables.
We carried out an experimental session based on four regression datasets
publicly available \citep{regressiondatasets}. For computational reasons,
we have limited the number of samples per dataset to 600 (randomly
sampled). The name of the datasets together with the number of variables
and number of samples are reported in Table (\ref{tab:Datasets}). 

\begin{table}
\begin{centering}
\begin{tabular}{|c|c|c|c|}
\hline 
dataset & name & $n$ & $m$\tabularnewline
\hline 
\hline 
1 & Ailerons & 35 & 600 \tabularnewline
\hline 
2 & Pol & 26 & 600 \tabularnewline
\hline 
3 & Triazines & 58 & 186 \tabularnewline
\hline 
4 & Wisconsin & 32 & 194\tabularnewline
\hline 
\end{tabular}
\par\end{centering}

\caption{Regression datasets, together with their number of variables $n$
and number of samples $m$, used as benchmark.\label{tab:Datasets}}
\end{table}

In order to eliminate a possible variable selection bias, each dataset
is first divided into two equal parts, one for ranking variables and
one for evaluating those rankings. The evaluation
of a ranking method is given by the mean squared error returned by
a 10-fold cross-validation of a linear regression (R lm function),
a SVM with radial kernel (R package e1071) and a random forest (R
package randomForest). In order to avoid the bias related to the size
of the feature set, we average the performance over all the feature
sets size (that range from 2 to 10 for each dataset) \citep{bontempiICML2010}.
 Table (\ref{tab:real-data-1}) reports the wins and losses on the
four datasets for each learning algorithm as well as per datasets.

\begin{table}
\begin{centering}
\begin{tabular}{|c|c|c|}
\hline 
rank-$\iota^{2}$ vs rank-$\rho^{2}$ & wins & loss\tabularnewline
\hline 
\hline 
Linear model & 2 & 2\tabularnewline
\hline 
Random forest & 2 & 2\tabularnewline
\hline 
SVM radial & 2 & 2\tabularnewline
\hline 
total & 6 & 6\tabularnewline
\hline 
\end{tabular}\\

\begin{tabular}{|c|c|c|}
\hline 
rank-$\max\iota^{2}$ vs rank-$\rho^{2}$ & wins & loss\tabularnewline
\hline 
\hline 
Ailerons & 0 & 3\tabularnewline
\hline 
Pol & 0 & 3\tabularnewline
\hline 
Triazines & 3 & 0\tabularnewline
\hline 
Wisconsin & 3 & 0\tabularnewline
\hline 
\end{tabular}
\par\end{centering}

\caption{Comparison of ranking strategies based on $\rho^{2}(X,Y)$ and $\max\iota^{2}(X,Y)$.
Wins and losses are defined by the lower 10-fold-cross-validated mean
squared error returned by a linear regression, a radial SVM and a
random forest averaged over subset sizes ranging from 2 to 10, for
each of the four datasets. \label{tab:real-data-1}}
\end{table}

We observe here that $\max\iota^{2}$ outperform $\rho^{2}$ on two
datasets and underperform $\rho^{2}$ on the two others, those results
are independent of the learning strategy used.

\section{Conclusion}

The goal of this paper has been to introduce a new measure of bivariate
dependency called a minrelation. We defined a new statistical rank
coefficient to determine if two continuous variables are minrelated
(or respectively majrelated). Finally, we showed the usefulness of
the minrelation coefficient on toy examples as well as on artificial
and real datasets. Indeed, $\max\iota^{2}$ appears to be a competitive
criterion w.r.t. Spearman's $\rho^{2}$ for ranking variables. We
deem that competitive results with Spearman's correlation makes
this coefficient an appealing new tool for the toolbox of any data
analyst. Furthermore, we believe that specific selection strategies
that take into account the directionality of the minrelation will
hold bigger promises. However, further research should focus on that
topic as well as on the limitations of this new measure (i.e. sample
statistic, linearity assumptions or even spurious case of high iota
values).

\end{document}